\pdfoutput=1

\documentclass[a4paper, 10pt, conference]{ieeeconf}      

\IEEEoverridecommandlockouts                              

\usepackage{epsfig} 
\usepackage{ tipa }
\usepackage{ textcomp }
\usepackage{float}
\usepackage{ dsfont }
\usepackage[ruled,vlined,noresetcount]{algorithm2e}
\usepackage[noend]{algpseudocode}
\usepackage[
font=footnotesize 
]{subfig}
\usepackage{cite}
\usepackage{amsmath,amssymb,amsfonts}
\usepackage{graphicx}
\usepackage{textcomp}
\usepackage{xcolor}
\usepackage{url}
\SetKwFunction{KwFn}{Fn}
\SetKwProg{Fn}{Function}{}{}

\title{\LARGE \bf
PCR-Pro: 3D Sparse and Different Scale Point Clouds Registration and Robust Estimation of Information Matrix For Pose Graph SLAM
}

\author{M. Usman Maqbool Bhutta$^{1}$ and Ming Liu$^{2}$
\thanks{*This paper is supported by Shenzhen Science, Technology and Innovation Commission (SZSTI) JCYJ20160428154842603 also supported by the Research Grant Council of Hong Kong SAR Government, China, under Project No. 11210017 and No. 16212815 and No. 21202816, NSFC U1713211 awarded to Prof. Ming Liu}
\thanks{$^{1}$BHUTTA, M. Usman Maqbool is a Ph.D. candidate at Robotics and Multi-Perception Lab, Department of Electronic and Computer Engineering,
        The Hong Kong University of Science and Technology, HK
        {\tt\small mumbhutta@ust.hk}}%
\thanks{$^{2}$Ming Liu is an Assistant Professor at Department of Electronics and Computer Engineering and also serves as Director of Robotics and Multi-Perception Lab,
         The Hong Kong University of Science and Technology, HK
        {\tt\small eelium@ust.hk}}%
}

\begin{document}

\maketitle
\thispagestyle{empty}
\pagestyle{empty}

\begin{abstract}
	For both indoor and outdoor environments, we propose an efficient and novel method for different scales and sparse 3D point clouds registration that cannot be handled by the current popular ICP approaches. Our algorithm efficiently detects the scale difference between point clouds and uses the keyframes to estimate the relative pose for calculating the scale difference. The algorithm applies a filter and computes the final transformation which coverages to a global minimum. The good estimation of transform and scale helps in the calculation of the covariance matrix using a closed form solution efficiently. This covariance between point clouds helps in the estimation of information matrix for pose-graph SLAM.
\end{abstract}

\section{Introduction}
In the last decade, simultaneous localization and mapping (SLAM) has received a large amount of intention from researchers all over the world due to it vast applications in self-driving cars and indoor/outdoor autonomous robotics employment. In SLAM, the graph-based technique \cite{Grisetti2010} has become a very popular choice due to its robustness and efficiency for large-scale robotics purposes. The main goal of these graph-based approaches is to set up the state variables or parameters that maximally characterizes the effect of Gaussian noise on measurements, where robots poses are used in the optimization and minimization of the error function. In graph-based SLAM, the state variables are the state of the robot and position of the landmarks. These parameters can be estimated with the sensors of the robot. Therefore, the measurement of state variables depends only on the relative poses of the robots. For instance, in visual odometry, measurement depends only on the connected poses. \\
\begin{figure}[!h]
	\centering
	\includegraphics[width=.9\linewidth]{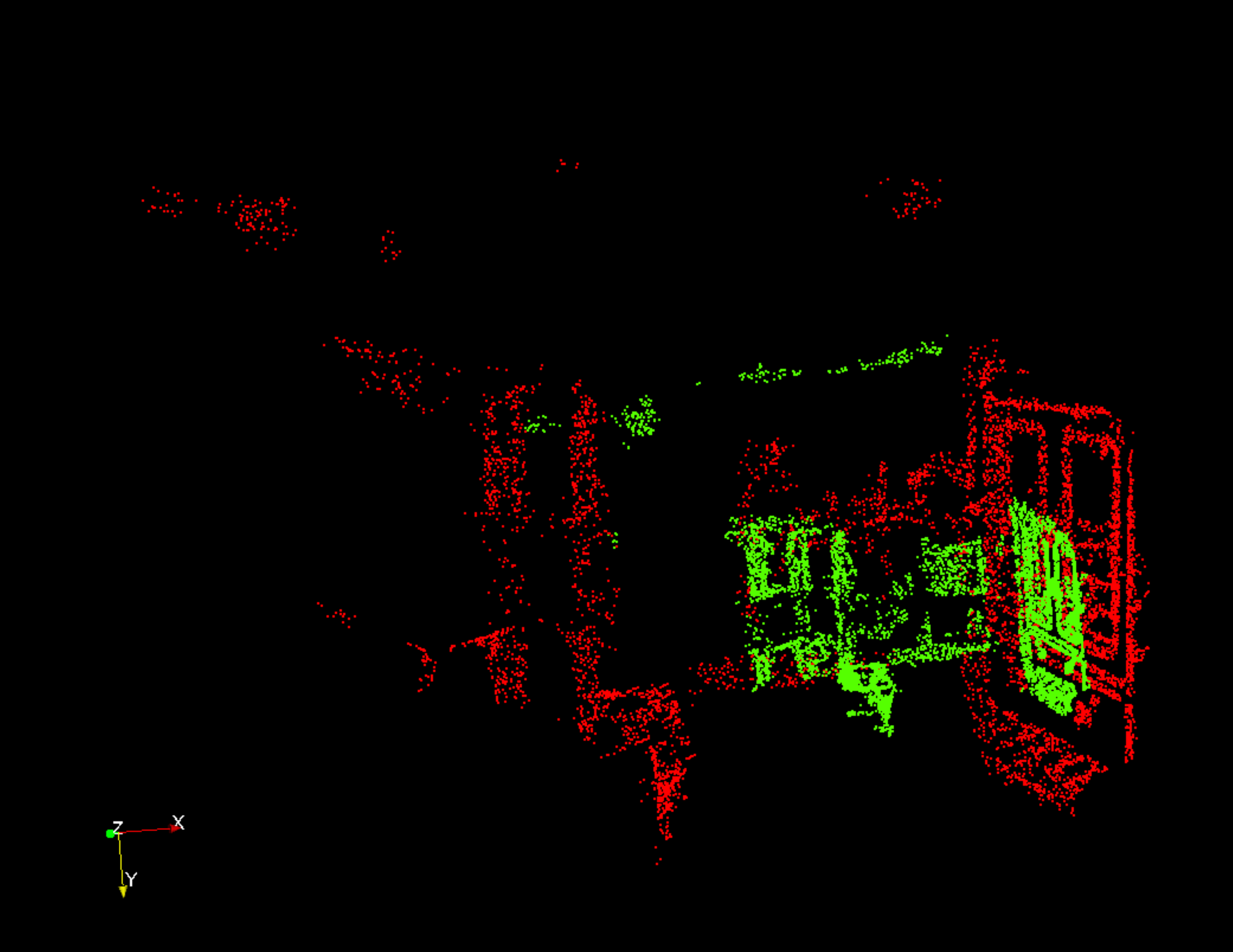}
	\caption{Scenario of the registration of sparse point clouds with scale difference. The green point cloud is the source point cloud while the target point cloud is shown in red.}
	\label{fig:1}
\end{figure}
In graph SLAM, state variables are defined as a node in the graph for optimization, and the pairwise information between two nodes defines the edge. After estimating the edge information, many algorithms have been described in the literature for further solving such a problem. A simple implementation can be done using some standard approaches such as Gauss-Seidel relaxation, Gauss-Newton and Levenberg-Marquardt (LM), which typically provide meaningful results for many applications. For problems that can be described as a graph, such as nonlinear least square problems, a general framework \textg $^{2}$o \cite{Kummerle2011} is used for optimization. This algorithm produces excellent results because it can exploit the sparsity between adjacent nodes of the graph and uses robust methods to solve the sparse linear system. \\
Given two point clouds with no scale difference, we can easily apply ICP to calculate the transformation, which basically maps all the points data of the source point cloud to the target point cloud. For aligning point clouds, multiple methods have been proposed \cite{Manoj2015,Pomerleau2011,doi:10.1177/0278364912458814, liu12robio}. Once we have a large point cloud map, we can use it in different applications, such as in autonomous navigation \cite{7110581, liu2014icra} and apply processing techniques such as segmentation within a point cloud \cite{Liu14robio, 8324799, liu2013icia}.\\

To find the edge information between two poses without any external sensors, such as an IMU, or wheels rotatory information is a challenging task in large-scale 3D mapping. Another difficult task is to find out the scale difference when two or more graphs are totally separated and have been created by different cameras and approaches, as shown in Fig.~\ref{fig:1}. Current ICP techniques are not sufficient to deal with such kinds of target problems. As multi-agent SLAM is the next interest for computer vision researchers, this creates a need for such a framework that actually accepts mapping information from different approaches and camera systems. For this scenario, point cloud registration of different scales is required. We thus propose a framework which will help deal with such kinds of problems. We strongly believe that it will help the computer vision community in many multi-agent scenarios. Our contribution to this framework is as follows;
\begin{itemize}
	\item It introduces a way of using direct SLAM approaches for multi-agent SLAM systems where we do not have information of the scale difference.
	\item It welcomes the features data to the featureless SLAM approaches.
	\item It presents a robust method to find an accurate transformation between sparse point clouds where the system converges to a global minimum.
	\item The covariance is very small, which will help in the usage of the information matrix and will solve many existing problems.
\end{itemize}  
\section{Overview of existing work}
For large scale 3D reconstruction difficulties, point cloud registration is a very popular choice.  ICP is a general and common approach to point cloud registration, which works interactively to find out pairs of closed points in a point clouds scan. The affine transformation (4x4) $T$ is calculated between two point clouds. This iterative algorithm minimizes a loss function $f(p_x, p_y, \omega)$ which is define as,

\begin{equation}\label{icp_equation}
f(p_x, p_y, \omega) = \sum_{i=1}^{n} { \parallel T (a_i,(p_x,p_y,\omega)) -b_{\theta(i)} \parallel }^2,
\end{equation}

where $T$ shows the affine transformation from the source scan to the  target scan, $\omega$ is the rotation value, and the translation in $x$ and $y$ are defined by $p_x$ and $p_y$, respectively.  $\theta$ interprets the index value of a point in the target point cloud, which is adjacent to the index $i$ of the source point cloud. In a 2D case, the minimization of the error function is done in the $x$, $y$ direction, and the rotation angle of the source scan to the target scan is optimized \cite{Rusinkiewicz}. \\

For SLAM using laser data, \cite{Burguera2005} proposed an efficient algorithm that firstly pre-processes the point cloud by removing the remote points in the point cloud data. After that, an assignment is done by finding out pairs of the closest points. Following, rejection is done by removing the pairs that have a long distance (by choosing the short-length pairs), and then using equation \eqref{icp_equation}, and the loss function is calculated. After applying the ICP transformation to source point cloud $PC_s$, the variance (average distance) between the aligned points is then estimated. Sometimes this variance is multiplied by the identity matrix to calculate the information model \cite{Endres2014}.\\

A modular ICP chain has been developed by \cite{Pomerleau2011}. This approach is known as libpointmatcher, which is open source software for ICP estimation. In simple terms, it calculates the translation and rotation matrix between two scans by minimizing the error function. The 3D source and target scans are sometimes named reference and reading, respectively, and the main goal is to align the reading with the reference. Libpointmatcher firstly applies different filtration algorithms to both the reference and the reading 3D scans and then starts the iteration by correlating points of the reading and reference point clouds. In each iteration, it finds out the transformation that reduces the alignment error. \\ 

The system iterates from the data filters to the error minimizer until the transformation conditions are satisfied or the alignment converges to their global minimum. A transformation checker is used for controlling the number of iterations. For instance, the fixed number of iterations or minimum error function threshold information can be used by the transformation checker. It can be used multiple times like the data filters in a chained form. The key idea for providing an interface for each step is to allow different algorithms at any step for evaluating the impact of the ICP behavior.  \\

Point cloud registration is a point cloud aligning problem that is used to merge different point clouds to get a large scale map if some matches are found. The libpointmacher and 3D ICP closed form solution  \cite{Manoj2015} performed very well and produced excellent results. But these approaches use dense point clouds as data with the same scale ( taken from the depth camera or lidar scan). In  the 3D closed form solution, for extraction of uniform key points, SHOT \cite{Salti2014} feature descriptors were used, which estimate nearest neighbor matches of the uniform keypoints features. This makes the ICP slightly slow compared to libpointmacher. While using the \cite{Pomerleau2011} libpointmacher and \cite{Manoj2015} 3D ICP closed form solution on sparse point clouds for registration using their respective ICP techniques, both the algorithms tried to align both point clouds and were unable to converge to a global minimum.\\

\section{Framework of the Proposed Approach}

\begin{figure}[!h]
	\centering
	\includegraphics[width=0.8\linewidth]{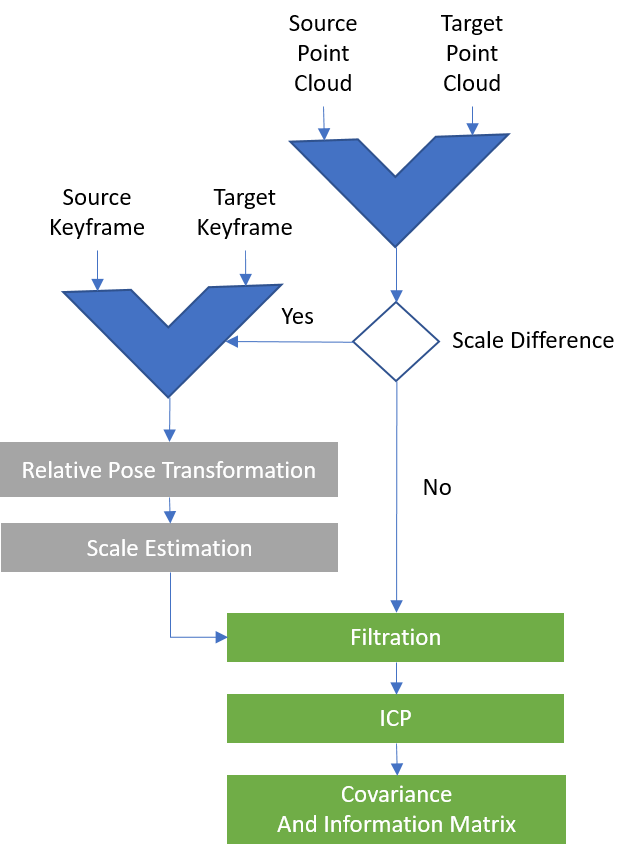}
	\caption{Overview of the proposed system (PCR-Pro) framework}
	\label{fig:11}
	\vspace{-1.5em}
\end{figure}

In SLAM using the camera, the point cloud is not as dense as it was when using lidar. Due to much fewer features in sparse point clouds, the chances for global convergence becomes very low using these methods. The scale problem is another factor, but our system (PCR-Pro) helps in the very efficient manipulation of sparse and different scale point clouds fusion. The framework of the PCR-Pro is shown in Fig.~\ref{fig:11}.\\

For estimating the robustness of the proposed algorithm, we consider the case of sparse points clouds, where the source 3D point cloud and target 3D point cloud are shown in green and red, respectively. The scale problem between them can be shown very clearly in Fig.~\ref{fig:1}.\\

\subsection{Scale Detection}

Our approach is to make the system work for both scenarios, whether they have a scale difference or they are almost equal. The first task is to find the difference using 3D sizes of both point clouds. If there is a difference detected in the volumes, then the system will first calculate the scale factor between them. \\
\subsection{Scale Estimation}
As discussed earlier, the scale plays a very important part in the alignment of point clouds. For the estimation of scale, 3D-point-clouds-related keyframes are used.\\

\begin{algorithm}
	\SetKwInput{KwData}{Input}
	\SetKwInput{KwResult}{Output}
	\KwData{keyframes, Point cloud of each keyframe}
	\KwResult{3D Transformation, Information Matrices }
	initialization\;
	\For{each point clouds}{detect scale difference} 
	\If{scale difference}{
		\For{each keyframe}{compute SIFT features and use FlannBased descriptor to find matches} 
		calculate matches between keyframes\;
		\For{all matches}{filter good matches} 
		\If{good matches}{
			by using opengv \cite{Kneip2014},  create RANSAC object\;
			set threshold\;
			set max iteration\;
			compute central relative pose $OT$ between keyframes    
		} 
		\For{all good matches}{compute scale difference $SC$ using kalman filter and relative pose $OT$}
		Transform source point cloud using scale transformation  $SC$\;
	}
	apply filter to crop lower area of source and target point clouds\;
	Estimate transformation $ T$ of filtered source and target point clouds using ICP \cite{Pomerleau2011}\;
	Transform original source point cloud using rigid transformation $RT $ will map on the target point cloud\;
	Now calculate of information matrix\;
	\Fn{3D ICP Covariance \cite{Rusu2011} (Source Point Cloud, Target Point Cloud, Final transformation $T$)}{
	 calculate the covariance using final transformation\;
	}
	 \Return 6x6 covariance matrix\;
	Information Matrix = (covariance matrix)$^{-1}$\;
	\caption{Proposed Approach}
\end{algorithm}

\subsubsection{Relative Pose estimation}

OpenGV \cite{Kneip2014} has a generalized description of sensors which makes it more reliable for central and non-central problems of the multi-camera system. It also supports estimation of absolute pose and relative pose for both central and non-central problems. It directly explores 3D information and estimation of the camera projection system using the image information without using the camera intrinsic and extrinsic properties. In relative pose estimation, OpenGV basically first creates the RANSAC object based on keypoint descriptors and their respective correspondences, and by using the threshold mentioned in equation \eqref{threshold} and maximum iteration information, it calculates the relative transformation:

\begin{equation}
\begin{aligned}
\epsilon_{threshold} = 1 - cos \theta_{threshold} = 1- cos(arctan \psi/l ),
\end{aligned}
\label{threshold}
\end{equation}

where $\psi$ is the classical reproduction error-threshold which is in pixels. The values are taken from the camera intrinsic parameters. \\
\\

\begin{figure}[!h]
	\centering
	\includegraphics[width=.9\linewidth]{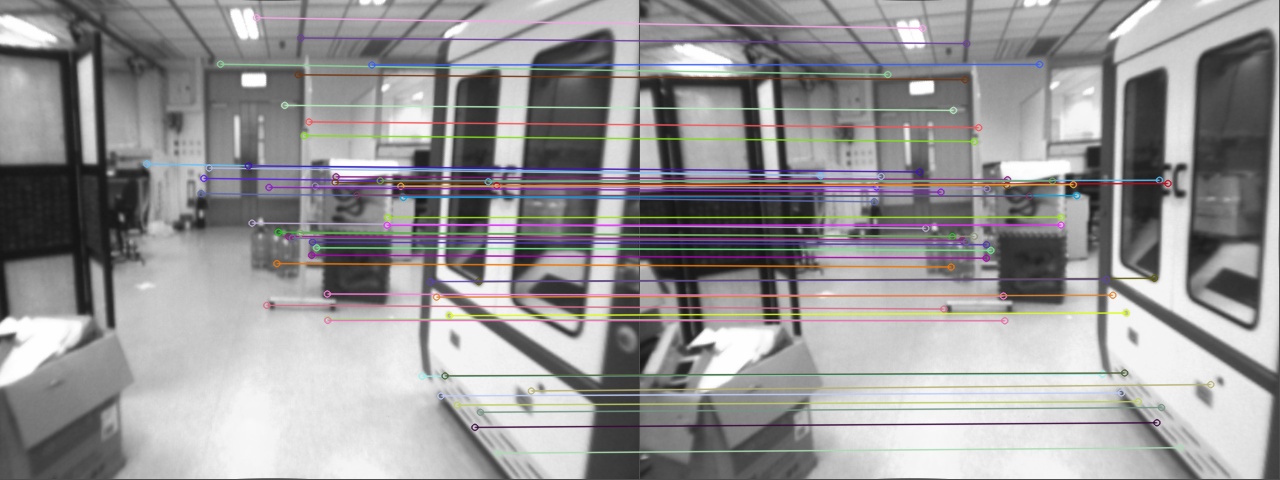}
	\caption{Feature matching between two keyframes}
	\label{fig:3}
\end{figure}


\subsubsection{Scale calculation using Kalman filter}
Camera intrinsic parameters are very important to find out the exact scale difference between two different keyframes of scenarios. These two keyframes can belong to two different cameras, but here we include the results of different sessions using the same camera. Firstly we calculate the feature matches between these keyframes,. We use a Flann-based descriptor \cite{muja_flann_2009} for SIFT feature matching, then filter good matches from all the feature matches, as shown in Fig.~\ref{fig:3}.  After filtering the good matches, the main goal is to find the transformation that can align both point clouds so that after applying the ICP, the system will converge to a global minimum, and minimize the covariance as well. \\
For all matches, we calculate the 2D points on the image, and then we map 2D points to 3D points using equation \eqref{eqn:3dpointfrom2d}.
\begin{equation}\label{eqn:3dpointfrom2d}
\begin{bmatrix} x \\ y \\ z \end{bmatrix}  = \begin{bmatrix} X * \frac{1}{f_x}+ \frac{-c_x}{f_x} \\Y * \frac{1}{f_y}+ \frac{-c_y}{f_y} \\ 1 \end{bmatrix},
\end{equation}
where ${X,Y} \in \mathds{R}^2 $, ${x, y, z} \in \mathds{R}^3 $ and $f_x,f_y,c_x $ and $c_y$ are taken from the camera intrinsic parameters.

Now that we have 3D points of the source and target keyframes, we apply relative pose transformation to all 3D points of the source keyframes. Using the Kalman filter, we increment the scale until all the transformed 3D points of the source keyframe are almost equal to the 3D points of the target keyframe. When the system converges to a global minimum, it will give us the scale difference, which we will then apply to their related point clouds.

\subsection{Filtration}
The source point cloud is transformed by the scale factor to remove the scale issue. Then both the source and target 3D scans are in almost the same scale form ready for the ICP for the indoor environment case. For the outdoor environment, a lot of noise is affected on the upper part of the point clouds where the depth is infinite (towards the sky). Therefore, there is a high error chance while in alignment and then applying the ICP. Our algorithm applies filtration to crop the lower 25\% of the point clouds. Most of the features are on the ground side which is sufficient for better alignments of both point clouds.\\

\subsection{Exact Transformation Computation}

\begin{figure}[!h]
	\centering
	\includegraphics[width=.9\linewidth]{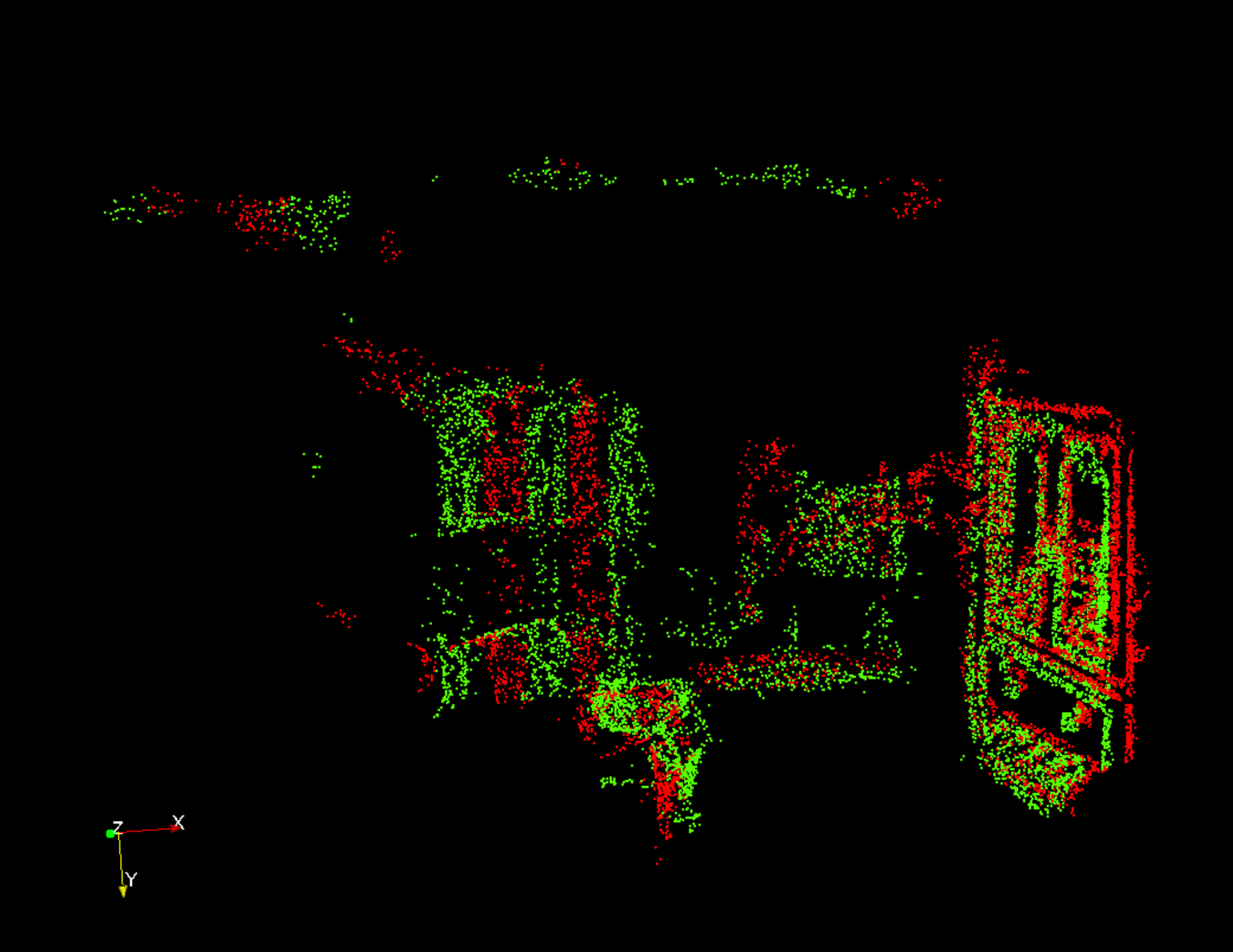}
	\caption{Source and target point clouds after scaling and registration using the proposed method}
	\label{fig:4}
\end{figure}
\begin{figure}[!h]
	\centering
	\includegraphics[width=.5\linewidth]{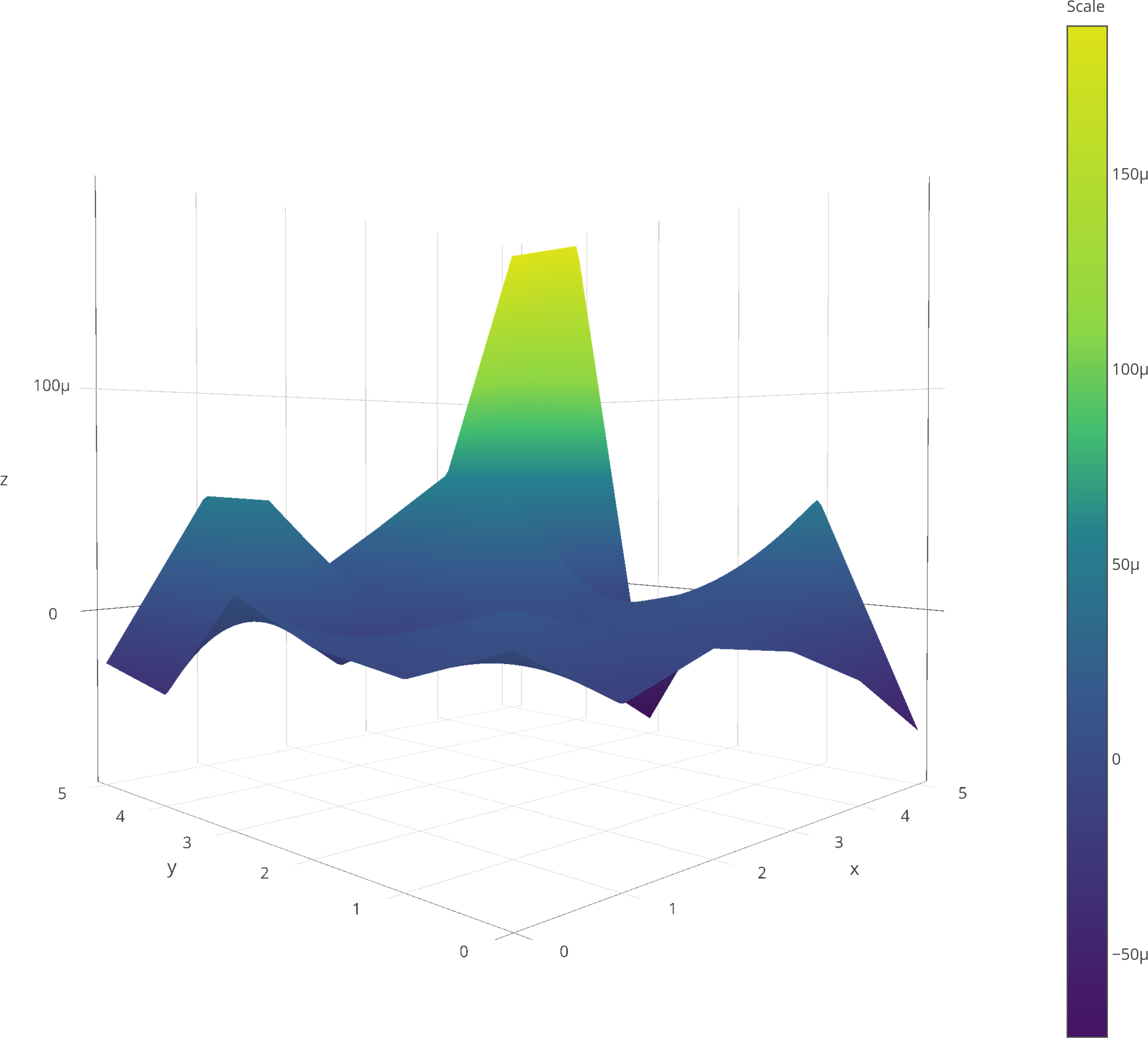}
	\caption{Covariance calculation}
	\label{fig:5}
\end{figure}

Using the filtered source and target point clouds, the libpointmatcher algorithm is applied to obtain the ICP transformation to get the global minimum. The system applies this transformation to the original source point cloud, which basically fuses with the target point clouds and gives excellent transformation, as shown in Fig.~\ref{fig:4}.\\

\subsection{Estimation of the information matrix}
For pose graph SLAM, optimization is done between two nodes by using the information matrix. The information matrix can be calculated by taking the inverse of the covariance matrix, which is a 6x6 matrix. The covariance matrix tells us about the covariance between two nodes and is usually, it is calculated from the sensors of robots. Recently  \cite{Manoj2015} developed a robust estimate of 3D ICP covariance that is a RANSAC-based approach. This method uses the concept of covariance as constraints that minimize the objective function \cite{Censi2007}.  \\
Firstly a closed-form expression for the Jacobian-based objective function is calculated, then the algorithm calculates the covariance using equation \eqref{2_eq}:
\begin{figure*}[!ht] 
	\centering
	\subfloat[Indoor environment II]{%
		\includegraphics[width=0.18\linewidth]{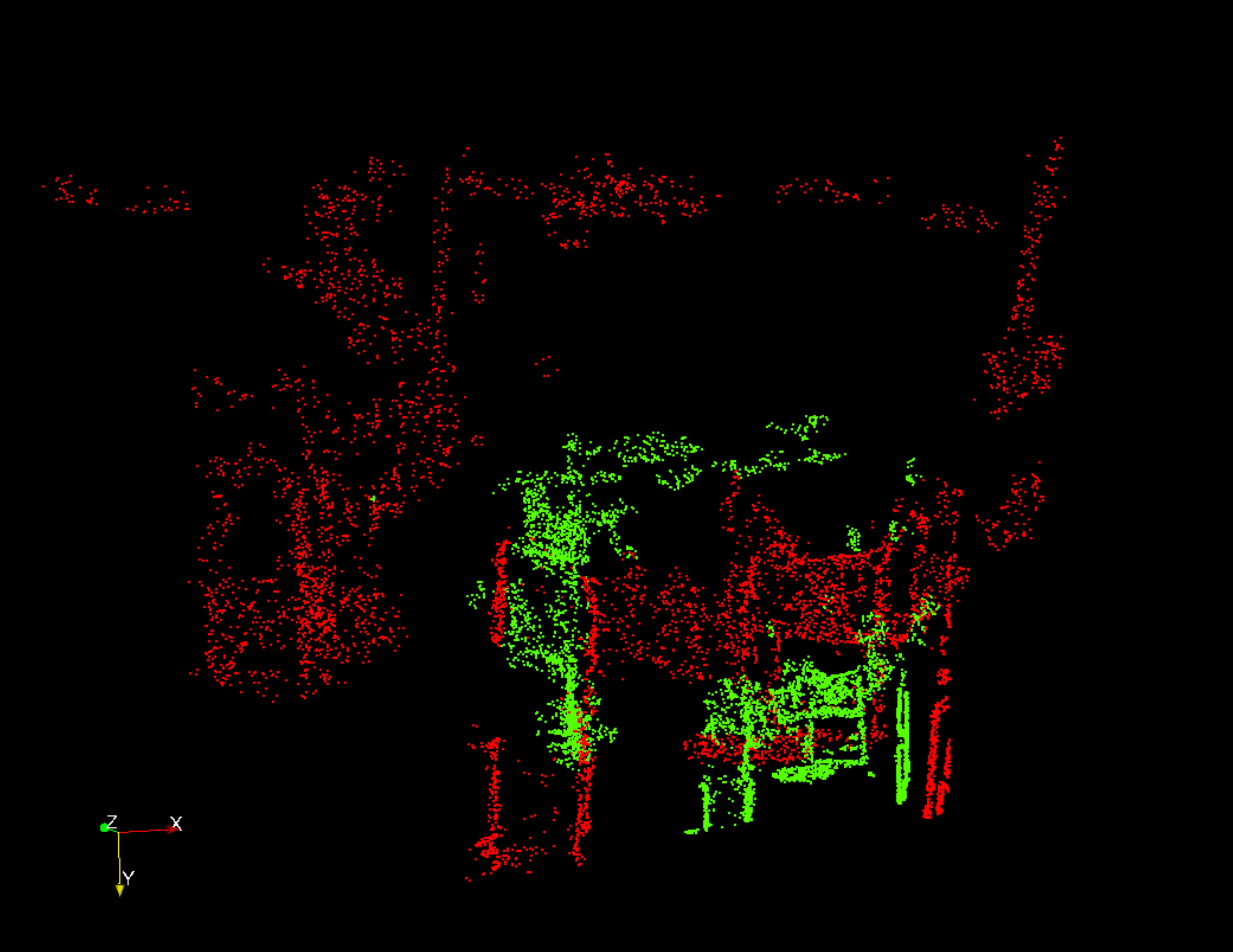}}
	\label{1a}
	\subfloat[Feature matching of its corresponding keyframes]{%
		\includegraphics[width=.36\linewidth]{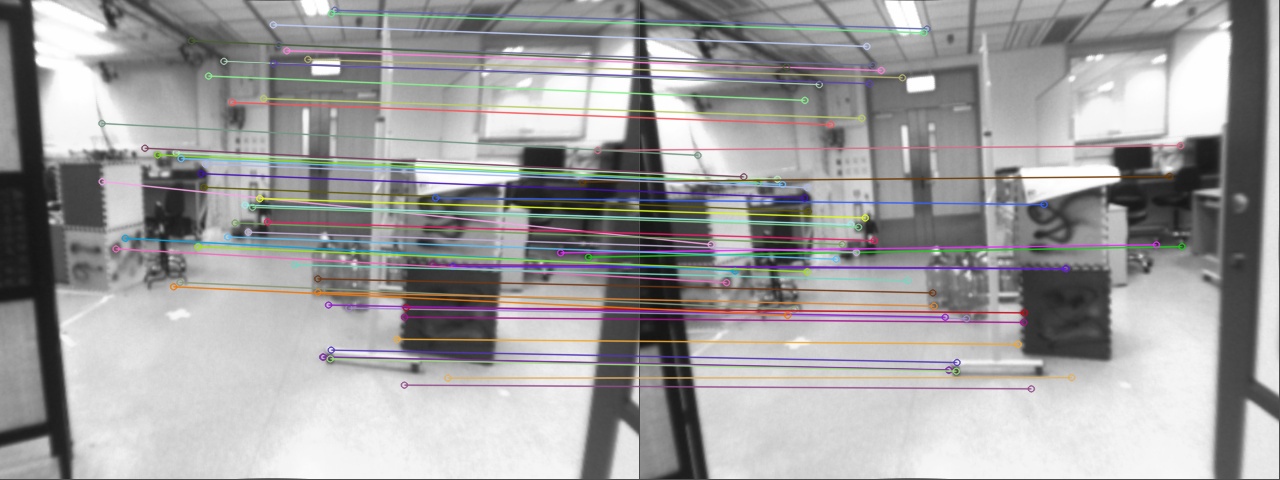}}
	\label{1b} 
	\subfloat[Point cloud fusion]{%
		\includegraphics[width=0.18\linewidth]{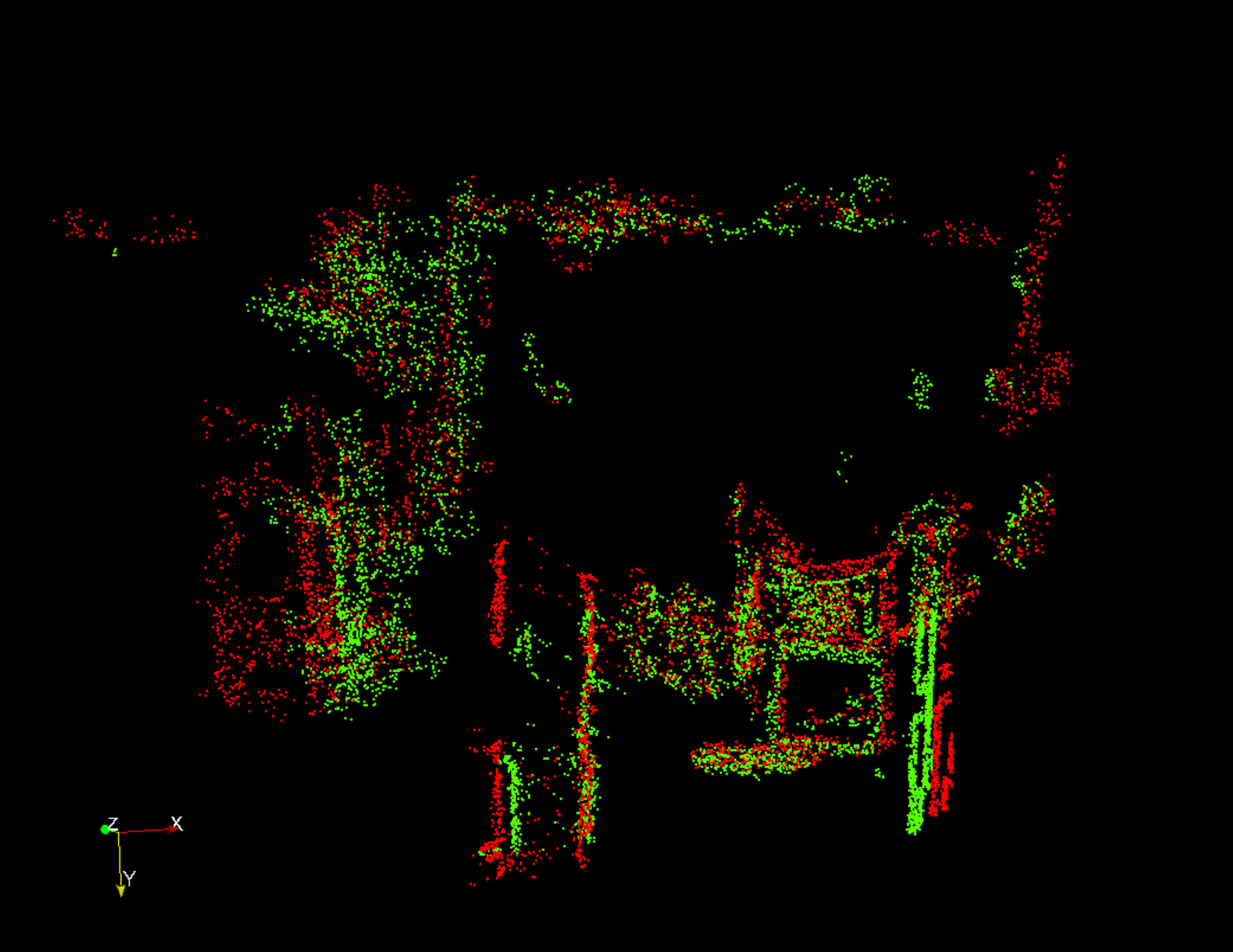}}
	\label{1c}
	\subfloat[Covariance after registration]{%
		\includegraphics[width=0.18\linewidth]{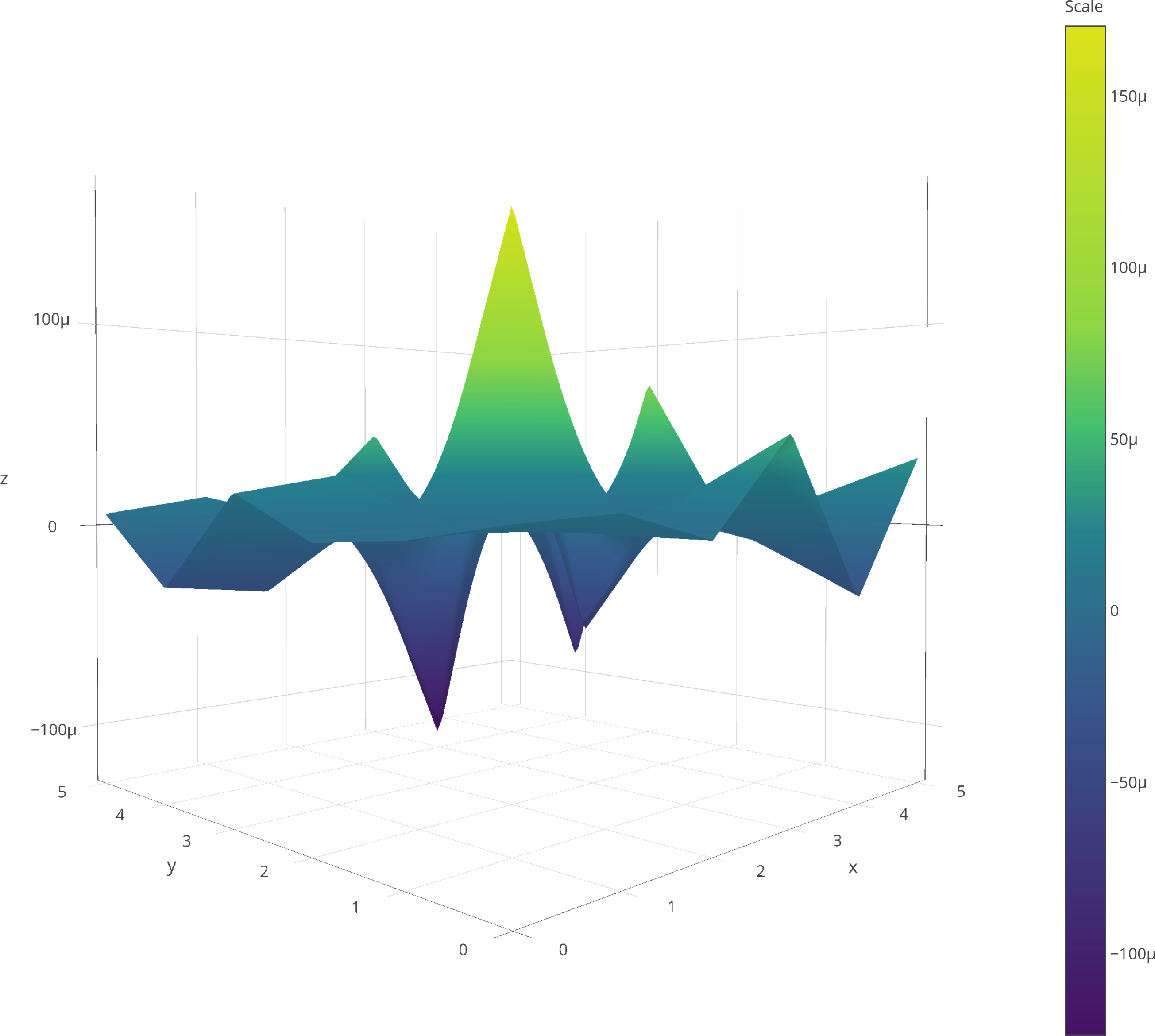}}
	\label{1d}\\
	\subfloat[Indoor environment III with
almost no scale difference]{%
		\includegraphics[width=0.18\linewidth]{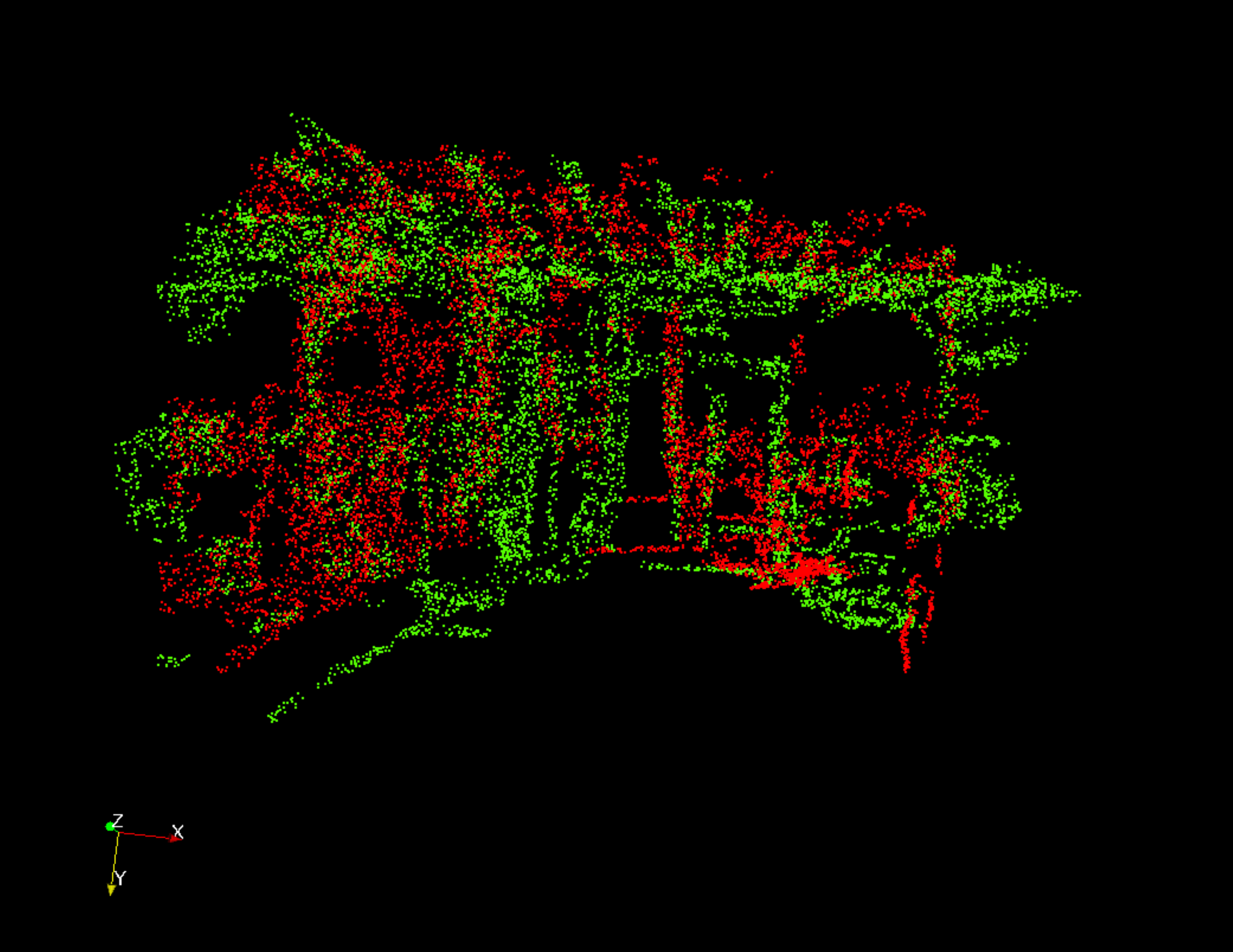}}
	\label{2a}
	\subfloat[Feature matching of its corresponding keyframes]{%
		\includegraphics[width=.36\linewidth]{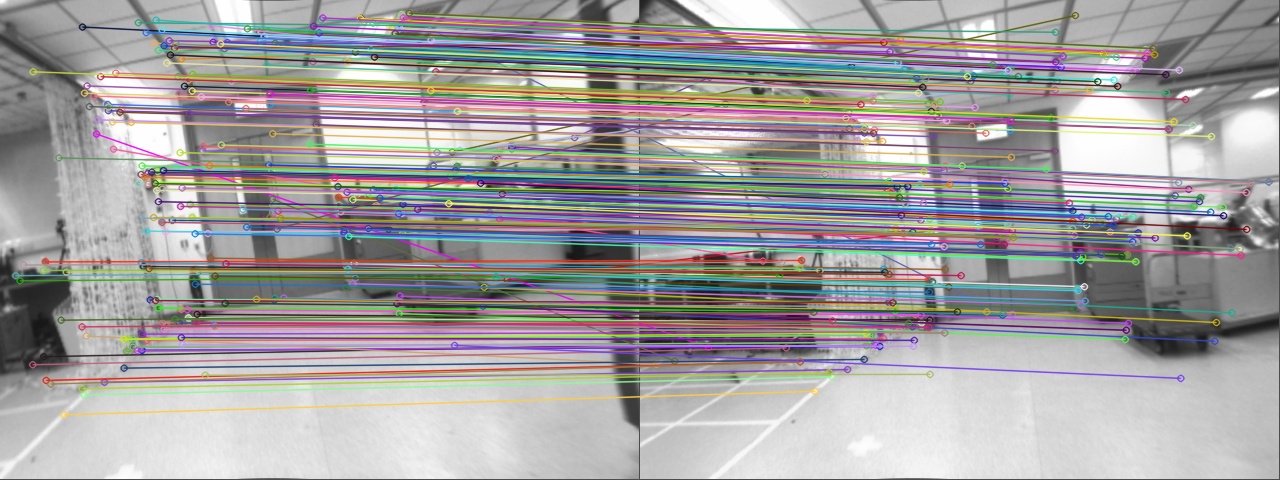}}
	\label{2b} 
	\subfloat[Point cloud fusion]{%
		\includegraphics[width=0.18\linewidth]{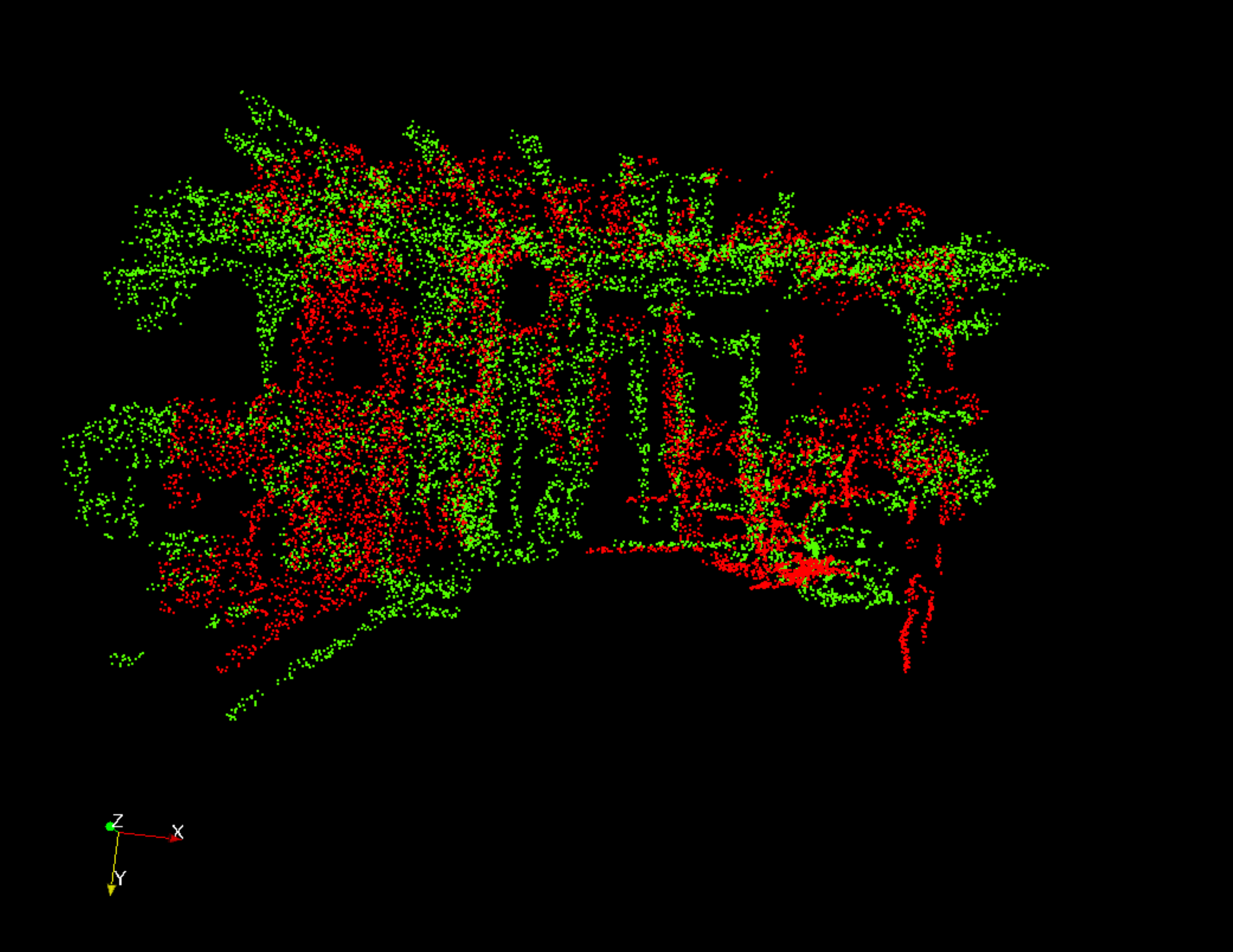}}
	\label{2c}
	\subfloat[Covariance after registration]{%
		\includegraphics[width=0.18\linewidth]{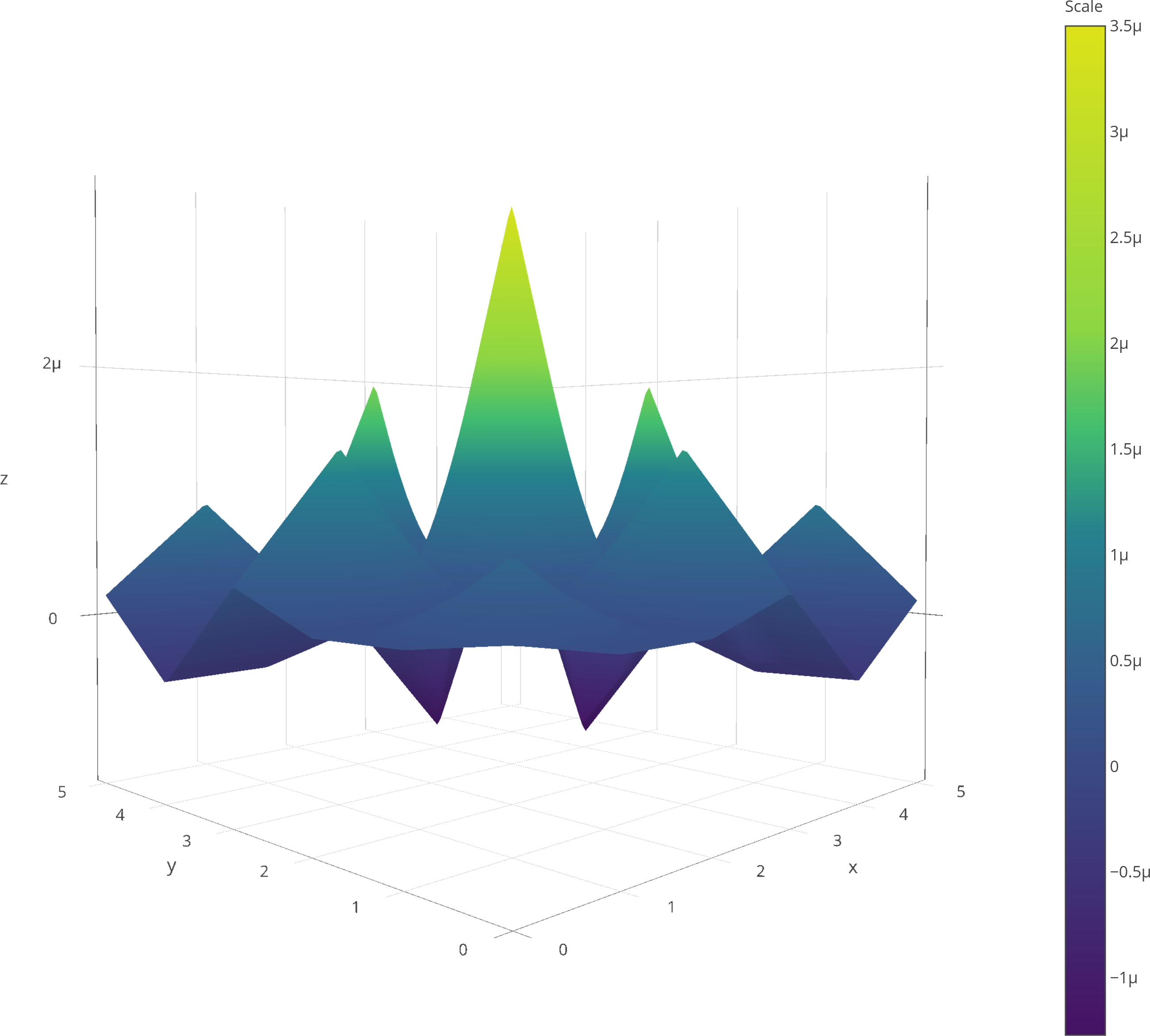}}
	\label{2d}\\
	\subfloat[Outdoor environment
I]{%
		\includegraphics[width=0.18\linewidth]{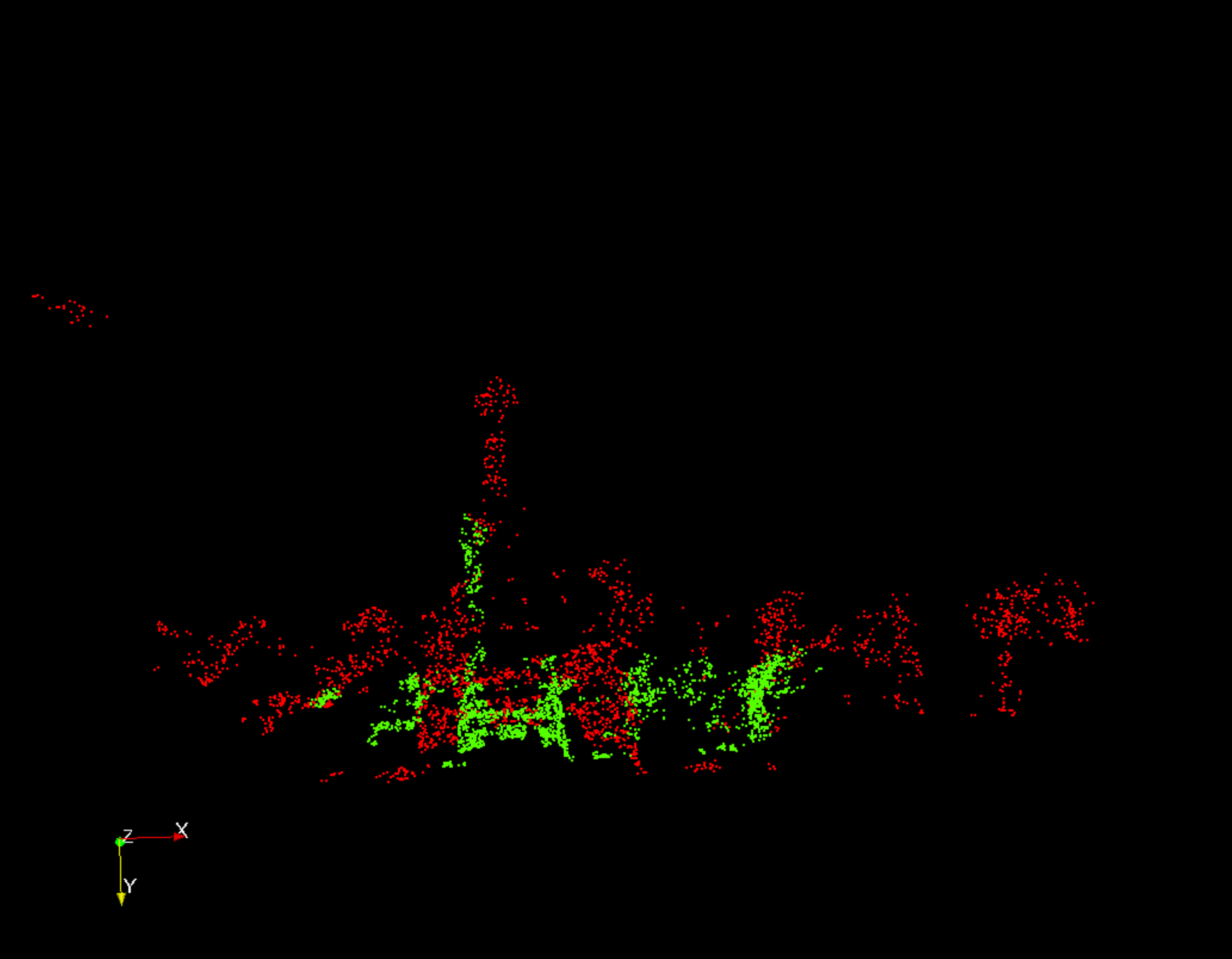}}
	\label{3a}
	\subfloat[Feature matching of its corresponding keyframes]{%
		\includegraphics[width=.36\linewidth]{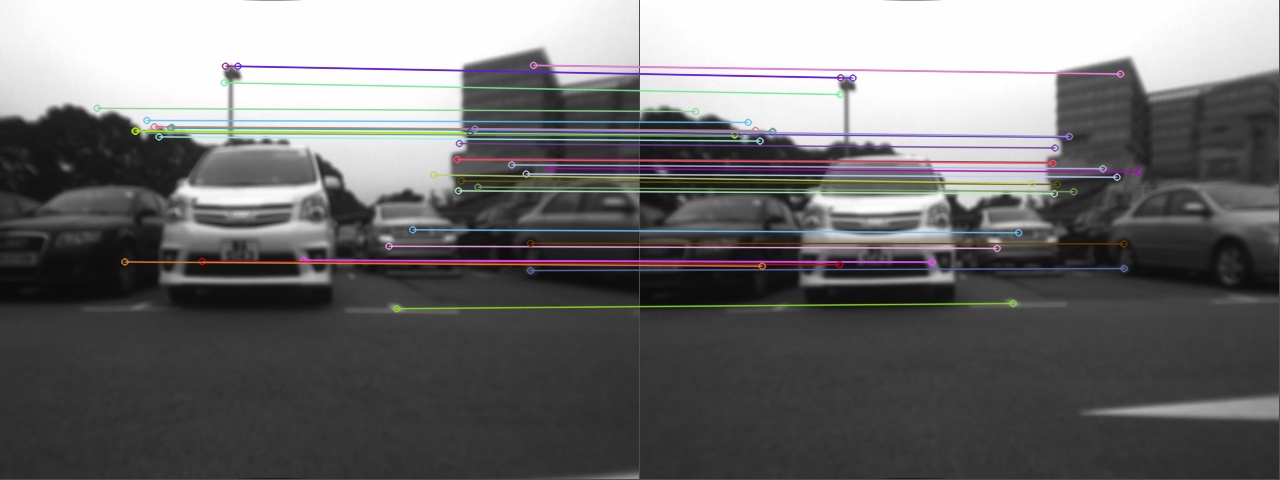}}
	\label{3b} 
	\subfloat[Point cloud fusion]{%
		\includegraphics[width=0.18\linewidth]{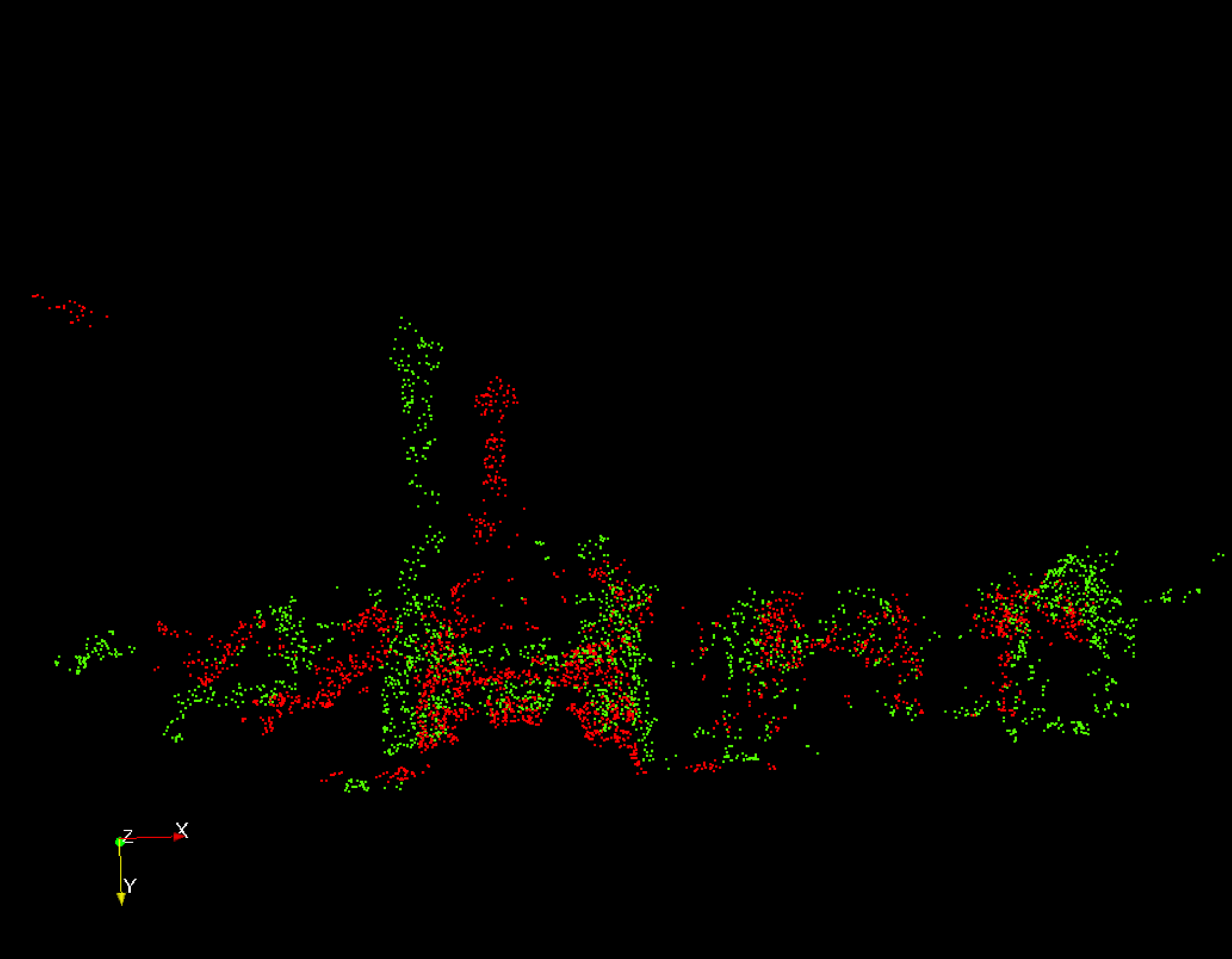}}
	\label{3c}
	\subfloat[Covariance after registration]{%
		\includegraphics[width=0.18\linewidth]{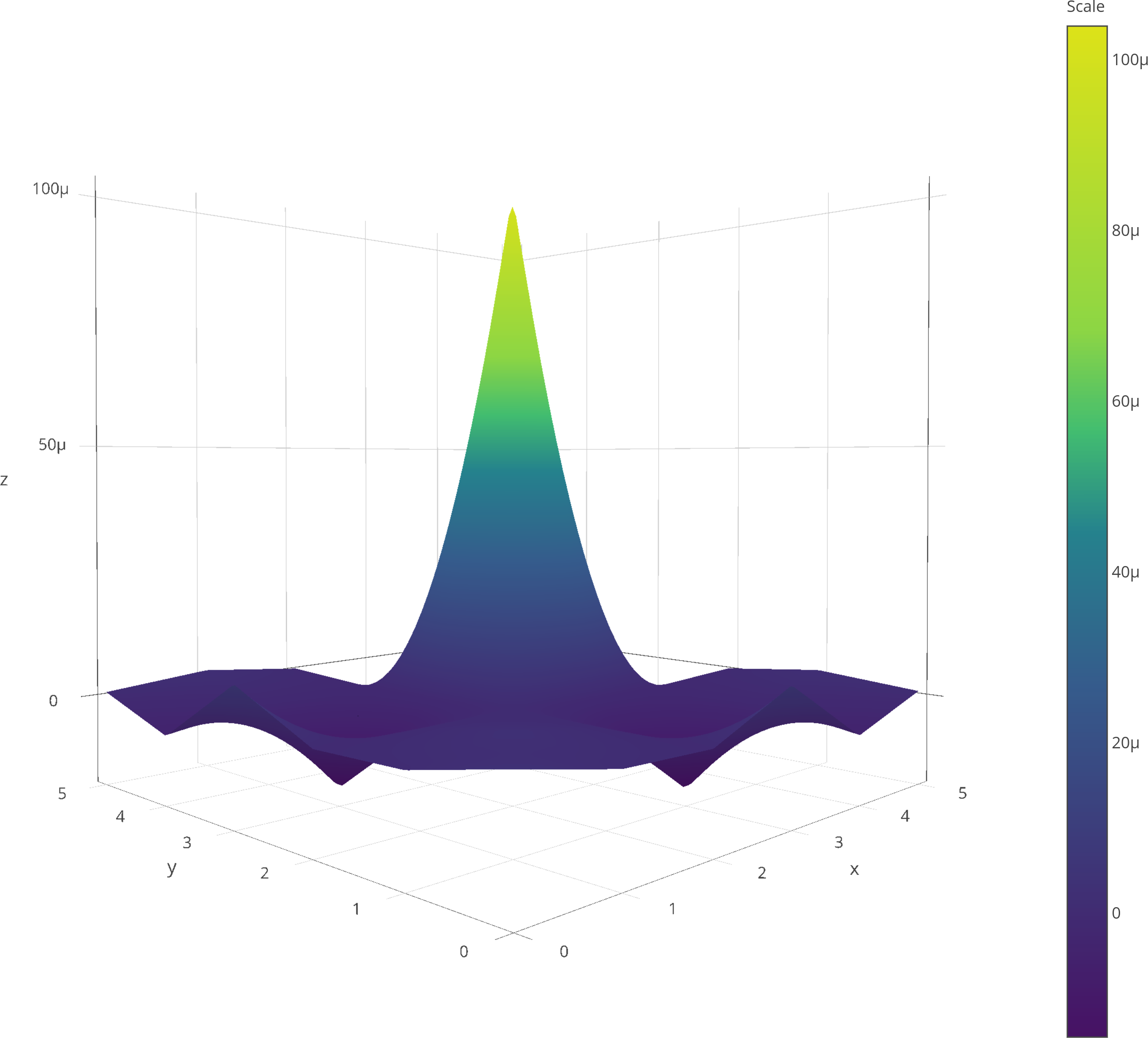}}
	\label{3d}\\
	\subfloat[Outdoor environment
II]{%
		\includegraphics[width=0.18\linewidth]{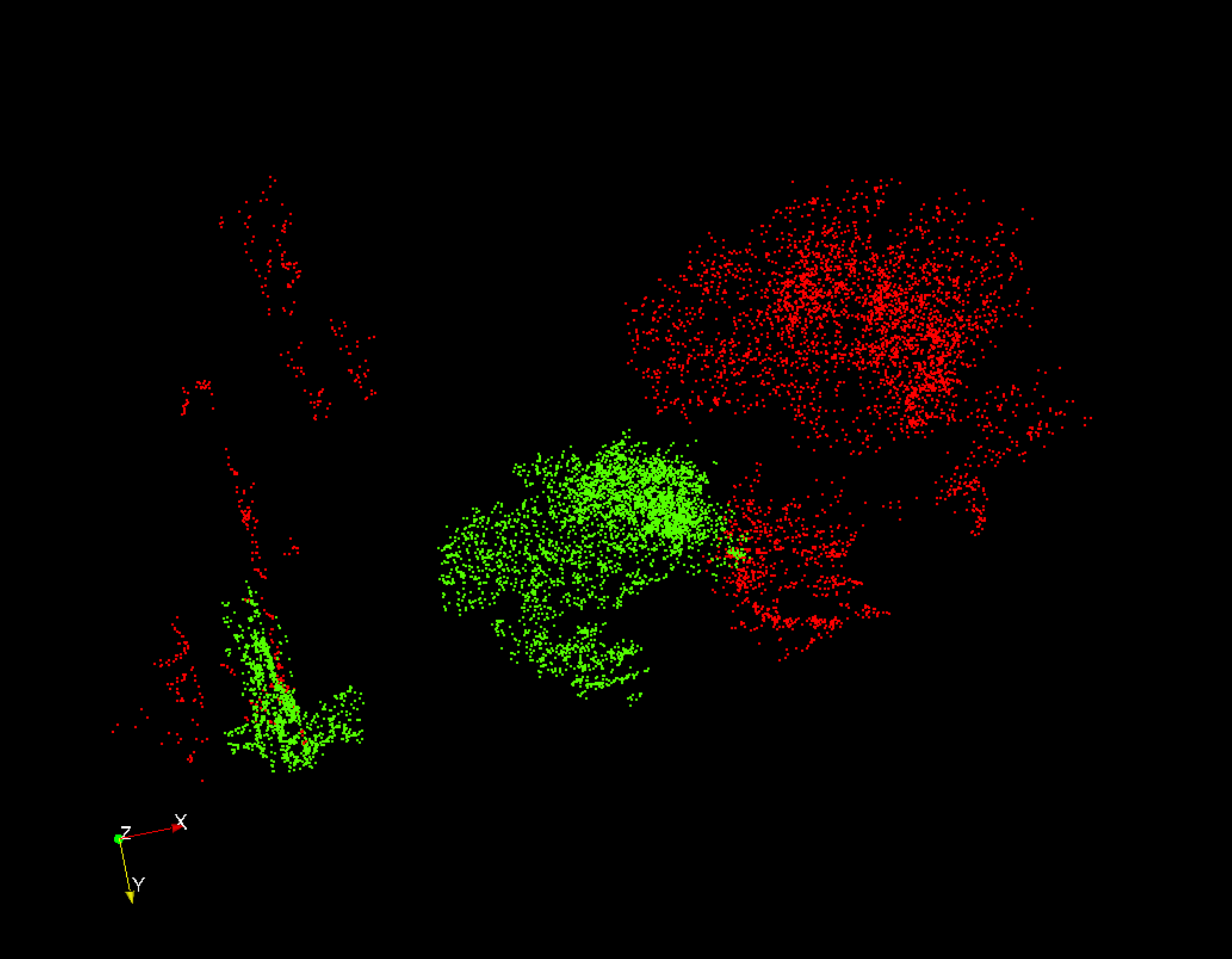}}
	\label{4a}
	\subfloat[Feature matching of its corresponding keyframes]{%
		\includegraphics[width=.36\linewidth]{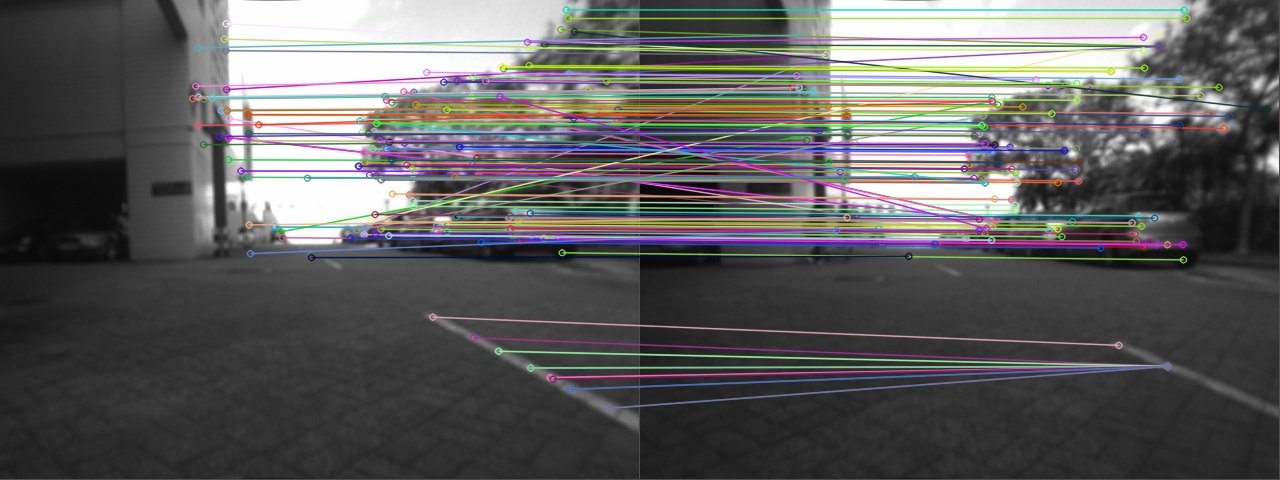}}
	\label{4b} 
	\subfloat[Point cloud fusion]{%
		\includegraphics[width=0.18\linewidth]{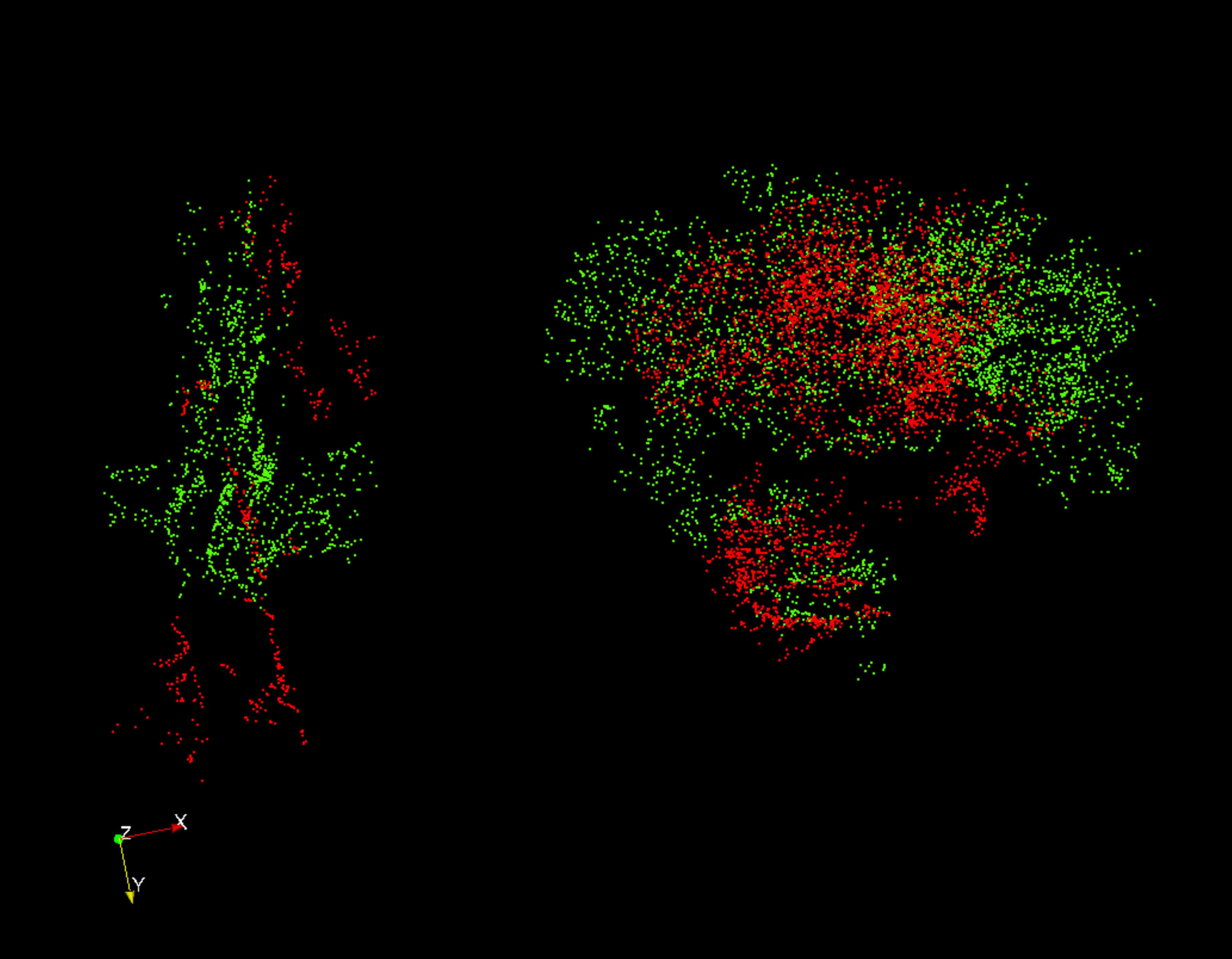}}
	\label{4c}
	\subfloat[Covariance after registration]{%
		\includegraphics[width=0.18\linewidth]{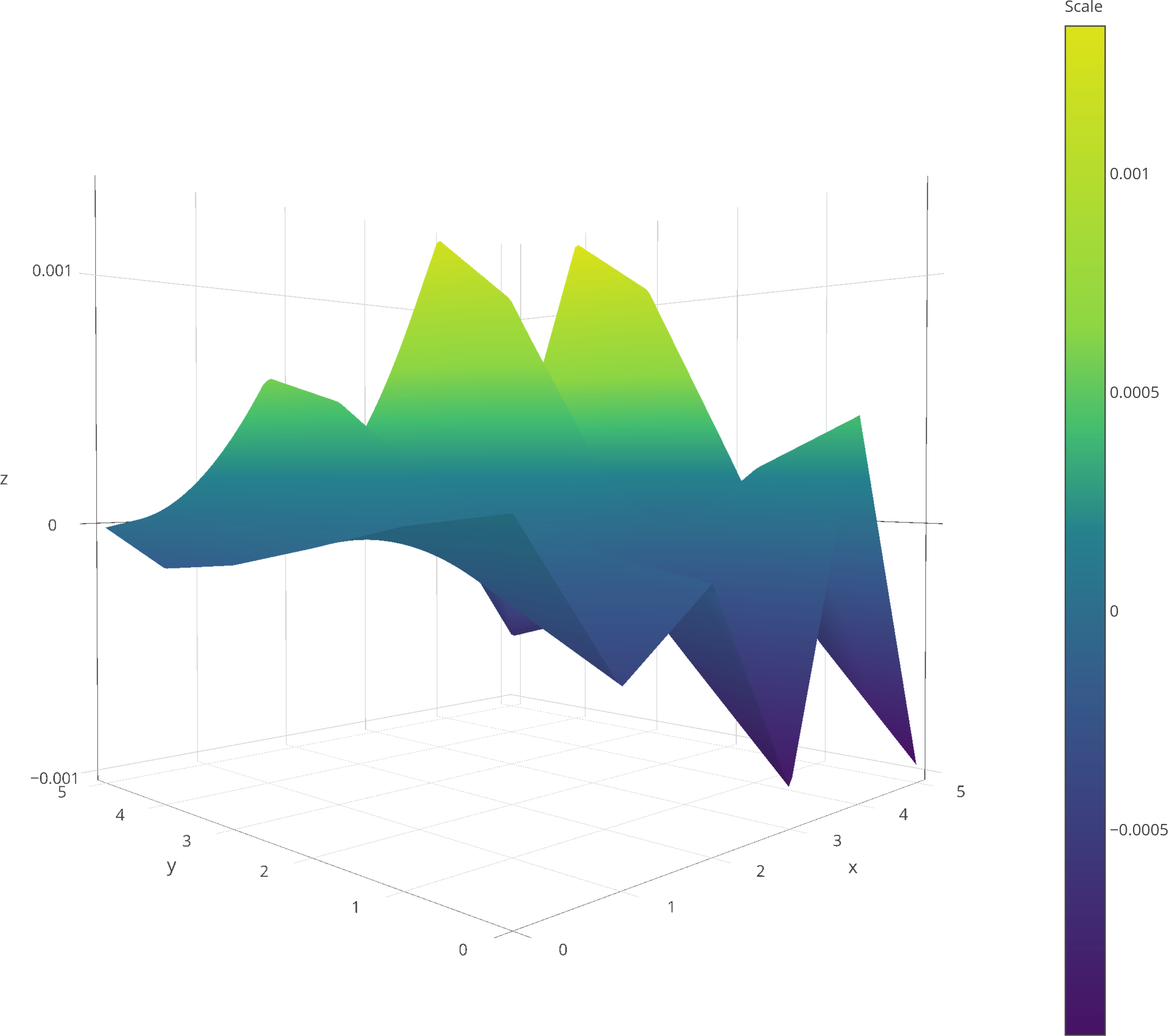}}
	\label{4d}\\
	\caption{Indoor and outdoor point cloud fusion and covariance results.}
	\label{fig:6} 
\end{figure*}
\begin{equation}\label{2_eq}
cov(x) \approx \left(\frac{\partial^2 J}{\partial x^2} \right)^{-1} \left( \frac{\partial^2 J}{\partial z\partial x} \right)cov(z) \left( \frac{\partial^2 J}{\partial z\partial x} \right)^T \left(\frac{\partial^2 J}{\partial x^2} \right)^{-1},
\end{equation}

In pose graph SLAM, for each pose, the keyframe transformation can be found out using OpenGV \cite{Kneip2014} between the two poses of the unknown robots (without knowing the camera matrix). The key idea is to find the covariance matrix by using the above closed-form method on the adjacent keyframes, and information for the pose graph SLAM.

For ICP, the problem can be written as
\begin{equation}
\begin{aligned}
J = {\text{minimize}}
\sum_{i=1}^{n}{ \parallel RP_i + T - Q_i  \parallel}^2.
\end{aligned}
\label{obj_funct}
\end{equation}
Equation \eqref{obj_funct} represents the objective function of the ICP for registration of point cloud $P$ and $Q$. $P_i$ and $Q_i$ show the 3D points. Furthermore, R and T are the rotation and translation, respectively, of homogeneous transformation H. The main goal is to estimate the covariance 6x6 matrix of the ICP transformation. Point to point matrix-based error function will be used, as mentioned in equation \eqref{obj_funct}.\\

Each set of correspondence $\{P_i, Q_i \}$  calculates $ \left(\frac{\partial^2 J}{\partial x^2} \right) $ in which $J_i$ depends on the value of the $i^{th}$ correspondence. After calculating each correspondence, it sums them to form the final estimate of $ \left(\frac{\partial^2 J}{\partial x^2} \right) $:
\begin{equation}
\begin{aligned}
\left(\frac{\partial^2 J}{\partial x^2} \right)   = \sum_{i=1}^{n} \left(\frac{\partial^2 J_i}{\partial x^2} \right).
\end{aligned}
\label{object_sum_corres}
\end{equation}

The author in \cite{Manoj2015} proposed a robust technique to calculate the covariance using the point clouds and can be calculated as;\\
\begin{equation}
\begin{aligned}
J = {\text{minimize}}
\sum_{i=1}^{n}{  F }^2,
\end{aligned}
\label{obj_funct_new}
\end{equation}
where $F = \parallel G \parallel$, and $G =  RP_i + T - Q_i$. \\

Our target is to find out the covariance using equation \eqref{2_eq}, which will use $J$. We efficiently calculate $J$ using equation \eqref{obj_funct_new}  and then successfully calculates the covariance matrix of size 6x6. As the algorithm converges to global minimum, so the resulting covariance is too small. The calculated values are shown in Fig.~\ref{fig:5}.  By taking the inverse transform, the information matrix is calculated. \\

\section{Numerical results and discussions}

Our main goal is to finalize a method for sparse point clouds registration to a global minimum. This also lessens the covariance between the correspondence of two 3D scans. The uncertainty in the transformation from one pose to another can be neglected if the transformation is correct, and an identity matrix can be used in that case while applying the pose graph optimization. \\
Using the feature matching technique, we aligned the source and target point clouds together. Then the PCR-Pro algorithm estimated the accurate transformation for point cloud registration, which minimized the point to point error between the correspondences (nearest neighbor points). Our algorithm produces excellent results and successfully converges to a global minimum, for both indoor and outdoor environments, and with or without scale difference, as shown in Fig.~\ref{fig:6}.\\
As the system converges  successfully for different sparse point clouds, as compared to other methods mentioned above, our estimated covariance is minimum. In pose graph SLAM, in this case, we can also take the identity matrix as the information matrix, and multiply it by a large number because the pose transformation is almost correct. Otherwise, we will estimate the covariance using equation \eqref{2_eq}. We have used LSD-SLAM \cite{Engel2014} for the collection of the dataset (keyframes and their respective point clouds) for validation of the algorithm. \\
After the fusion of point clouds in different indoor and outdoor cases, the covariance estimation results showed the robustness of PCR-Pro in all cases. Due to the best transformation and scaling for all scenarios, the estimated covariance is too small, as shown in Fig.~\ref{fig:6}. If the measurements from sensors $ z$ are uncorrelated with each other, then the covariance matrix will be diagonal. 
The information matrix is just an inverse of the 6x6 covariance matrix in 6DOF space, where these are the upper triangular values of the 6x6 information matrix, i.e., the inverse of the covariance matrix, representing the uncertainty between two nodes (edges).  \\
With an increase in the number of correspondences, the computation time to estimate the covariance also increases. Applying the certain threshold limit in the number of correspondences minimizes the computation time, which enhances the efficiency of the covariance estimation technique. Our proposed algorithm uses a C++ implementation and the source code can be accessed from \url{https://sites.google.com/view/pcr-pro}. \\
\section{Conclusions}

We discussed in detail the existing ICP algorithms and covariance estimation between two point clouds for pose graph SLAM. Our algorithm works excellently as compared to other ICP algorithms, regardless of the sparsity of the point cloud. Furthermore, the estimated covariance is also very small as the point cloud registration converges to the global optimal. For pose graph SLAM, estimation of the uncertainty using the covariance, and then calculation of the information matrix is quite difficult for a monocular camera SLAM system. We proposed an efficient algorithm that gives excellent results for transformation estimation, which can work for both real-time and off-time large-scale mapping applications.

\bibliographystyle{plain}

\end{document}